\documentclass[preprint,12pt]{elsarticle}

\usepackage[T1]{fontenc}
\usepackage{graphicx,verbatim}
\usepackage{array}
\usepackage{amssymb} 
\usepackage{multirow} 
\usepackage{makecell}
\usepackage{amsmath}
\usepackage{hyperref}

\journal{Computer Vision and Image Understanding}

\begin{document}

\begin{frontmatter}

\title{Challenging Vision-Language Models with Surgical Data: A New Dataset and Broad Benchmarking Study}


\author[dkfz]{Leon Mayer\corref{cor1}\fnref{projlead,core}}
\ead{leon.mayer@dkfz-heidelberg.de}
\author[dkfz,helmholtz]{Tim Rädsch\fnref{core}}
\author[dkfz,heidelberg-med,nct]{Dominik Michael}
\author[dkfz]{Lucas Luttner}
\author[dkfz,nct]{Amine Yamlahi}
\author[dkfz,nct]{Evangelia Christodoulou}
\author[dkfz,nct,hidss,heidelberg-cs]{Patrick Godau}
\author[dkfz,heidelberg-cs]{Marcel Knopp}
\author[dkfz,helmholtz]{Annika Reinke}
\author[purdue-bme,purdue-rche,indiana-bhs,indiana-surgery,dresden]{Fiona Kolbinger}
\author[dkfz,nct,helmholtz,heidelberg-med,heidelberg-cs]{Lena Maier-Hein}

\cortext[cor1]{Corresponding author}

\fntext[projlead]{Project lead}
\fntext[core]{Core contributor}

\affiliation[dkfz]{organization={German Cancer Research Center (DKFZ) Heidelberg, Div. Intelligent Medical Systems},
                  country={Germany}}

\affiliation[nct]{organization={National Center for Tumor Diseases (NCT), NCT Heidelberg},
                  country={Germany}}

\affiliation[helmholtz]{organization={DKFZ Heidelberg, Helmholtz Imaging},
                       country={Germany}}

\affiliation[hidss]{organization={HIDSS4Health - Helmholtz Information and Data Science School for Health},
                   country={Germany}}

\affiliation[heidelberg-med]{organization={Medical Faculty, Heidelberg University},
                            country={Germany}}

\affiliation[heidelberg-cs]{organization={Faculty of Mathematics and Computer Science, Heidelberg University},
                           country={Germany}}

\affiliation[purdue-bme]{organization={Weldon School of Biomedical Engineering, Purdue University},
                        city={West Lafayette},
                        state={IN},
                        country={USA}}

\affiliation[purdue-rche]{organization={Regenstrief Center for Healthcare Engineering (RCHE), Purdue University},
                         city={West Lafayette},
                         state={IN},
                         country={USA}}

\affiliation[indiana-bhs]{organization={Department of Biostatistics and Health Data Science, Richard M. Fairbanks School of Public Health, Indiana University School of Medicine},
                         city={Indianapolis},
                         state={IN},
                         country={USA}}

\affiliation[indiana-surgery]{organization={Department of Surgery, Indiana University School of Medicine},
                             city={Indianapolis},
                             state={IN},
                             country={USA}}

\affiliation[dresden]{organization={Department of Visceral, Thoracic and Vascular Surgery, University Hospital and Faculty of Medicine Carl Gustav Carus, TUD Dresden University of Technology},
                     city={Dresden},
                     country={Germany}}

\begin{abstract}
While traditional computer vision models have historically struggled to generalize to endoscopic domains, the emergence of foundation models has shown promising cross-domain performance. In this work, we present the first large-scale study assessing the capabilities of Vision Language Models (VLMs) for endoscopic tasks with a specific focus on laparoscopic surgery. Using a diverse set of state-of-the-art models, multiple surgical datasets, and extensive human reference annotations, we address three key research questions: (1) Can current VLMs solve basic perception tasks on surgical images? (2) Can they handle advanced frame-based endoscopic scene understanding tasks? and (3) How do specialized medical VLMs compare to generalist models in this context?
Our results reveal that VLMs can effectively perform basic surgical perception tasks, such as object counting and localization, with performance levels comparable to general domain tasks. However, their performance deteriorates significantly when the tasks require medical knowledge. Notably, we find that specialized medical VLMs currently underperform compared to generalist models across both basic and advanced surgical tasks, suggesting that they are not yet optimized for the complexity of surgical environments.
These findings highlight the need for further advancements to enable VLMs to handle the unique challenges posed by surgery. Overall, our work provides important insights for the development of next-generation endoscopic AI systems and identifies key areas for improvement in medical visual language models.

\end{abstract}

\begin{keyword}
Surgical Scene Understanding \sep Vision-Language Models \sep Endoscopic VQA Benchmarking
\end{keyword}

\end{frontmatter}

\section{Introduction}
In the context of computer vision, the endoscopic domain presents unique challenges from a computer vision perspective, with endoscopic images characterized by poor contrast and organs lacking sharp edges while overlapping substantially—features that diverge significantly from natural images. Consequently, methods developed for general computer vision have often failed to generalize well to surgical applications, necessitating the development of custom solutions such as for surgical phase recognition \cite{tecno_czempiel} and surgical action recognition \cite{selfdistill_yamlahi}.

With the emergence of foundation models (FMs), cross-domain generalization has improved dramatically, exemplified by DepthAnything \cite{yang2024depthanythingunleashingpower} and Segment Anything's \cite{sam_kirillov} successful application to endoscopic data \cite{liu2024surgicalsam2realtime,yuan2024segmentmodel2need}. However, the endoscopic domain remains relatively unexplored in the context of vision-language models. While first endoscopic vision-language solutions, such as HecVL \cite{hecvl_yuan} and VidLPRO \cite{vidlpro} have been proposed, they employ CLIP-style \cite{clip_radford} evaluation methods but lack open-ended visual question answering (VQA) capabilities.

This study poses the fundamental question: Do VLMs understand endoscopic images? In this context, we identified three critical gaps in the literature: First, while several endoscopic VQA datasets exist, the heterogeneity of questions prevents meaningful comparison with natural domain performance. Second, most endoscopic VQA studies utilize small, specific models without assessing state-of-the-art (SOTA) VLMs at scale. Third, no comparison exists between size-matched generalist and medical/surgical models on endoscopic tasks, despite evidence suggesting generalist models may outperform domain-specific ones.

\begin{figure}[t]
\includegraphics[width=\textwidth]{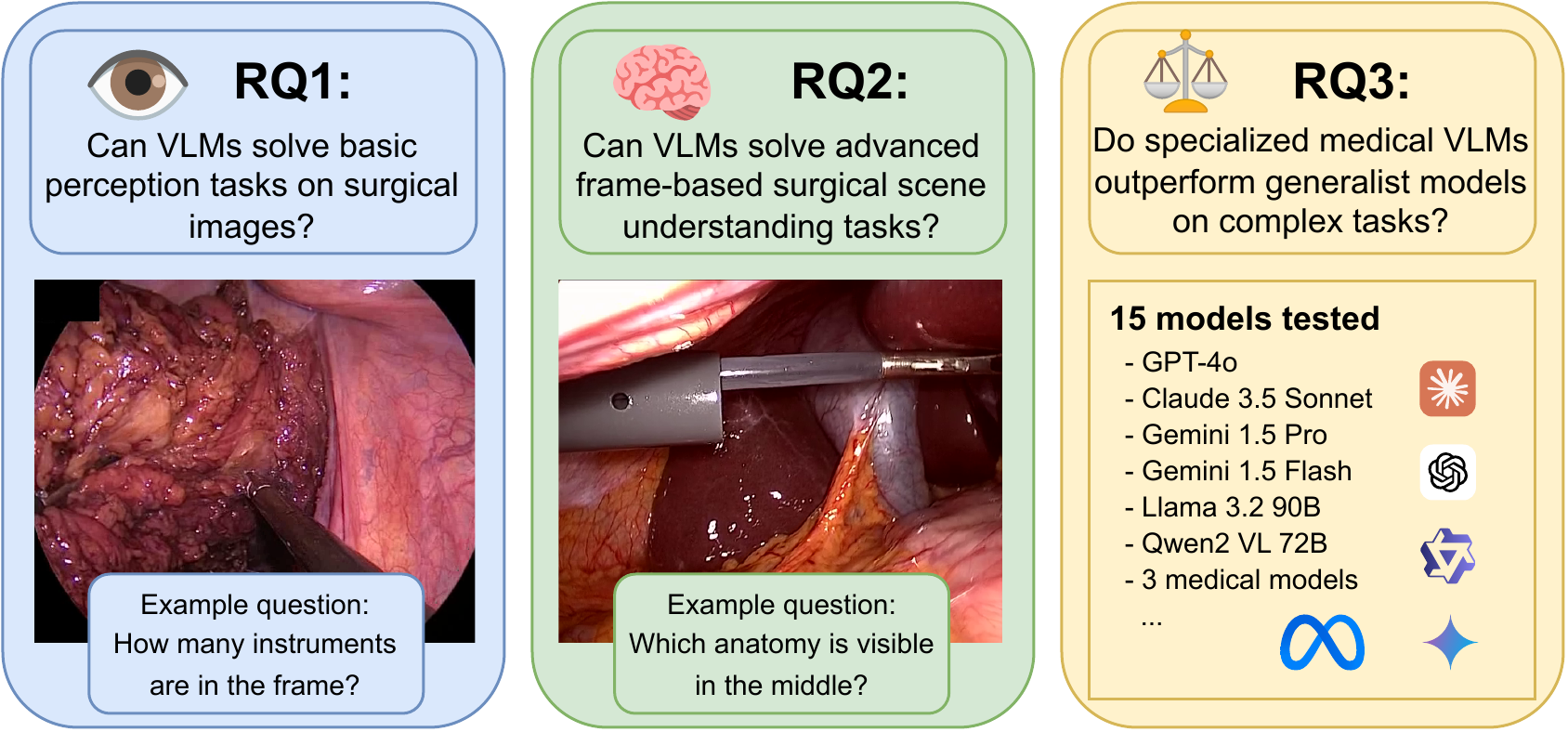}
\caption{\textbf{Research questions addressed in this study} and sample images illustrating the nature and complexity of basic perception and advanced tasks.} \label{figure:fig1}
\end{figure}

To address these gaps, we present a large-scale study investigating the three research questions depicted in Fig. \ref{figure:fig1}. Our work introduces the new public dataset HeiCo-VQA-Base comprising 24,252 images with 167,384 questions and, for 10\% of the questions, corresponding human baseline annotations.

\begin{figure}[t]
\includegraphics[width=\textwidth]{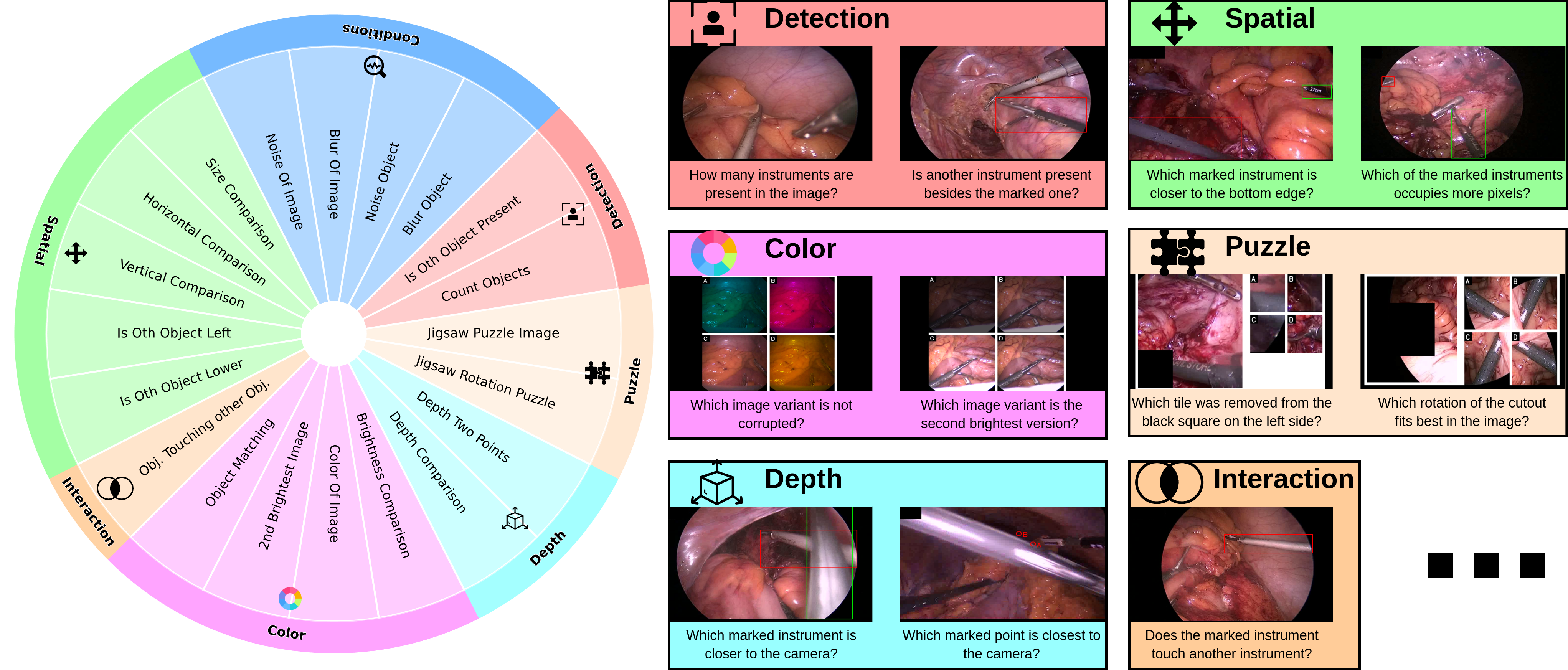}
\caption{\textbf{The basic perception tasks cover a broad range of question categories.} Example questions rephrased for brevity.} \label{figure:fig2}
\end{figure}

\section{Methods}
This section introduces our framework for VQA benchmarking (sec.  \ref{section:sec21}), the new surgical VQA benchmark we release as part of this work (sec. \ref{section:sec22}) as well as the specific experiments performed to address our core research questions (sec. \ref{section:sec23}).
\subsection{Framework for VQA Benchmarking} \label{section:sec21}
To investigate the RQs depicted in Fig. \ref{figure:fig1}, we developed a flexible framework that enables integration of any state-of-the-art VLM while assessing performance across varying levels of complexity.

\textbf{Basic perception tasks} evaluate a model’s fundamental visual understanding without requiring any medical knowledge, thereby enabling cross-domain comparability. We convert a dataset of instance segmentations into a VQA dataset using the following pipeline adopted from \cite{bridging_raedsch}. First, each image and its objects are enriched with relative depth information via the Depth Anything V2 model \cite{yang2024depthv2}. Next, a set of rules transforms object attributes, including depth, size, color, and spatial relationships of surgical instruments, into a set of facts (e.g., “there are N instruments in the image”). Following established VQA categorization \cite{blink_fu}, the resulting questions are subdivided into seven categories, depicted in Fig. \ref{figure:fig2}. All random baselines in our work are computed by selecting answers uniformly at random. 

\begin{table}[t]
\caption{\textbf{Overview of the utilized endoscopic datasets. }These cover a broad range from general perception tasks to very hard endoscopic tasks.}
\centering
\scriptsize
\renewcommand{\arraystretch}{1.2} 
\setlength{\tabcolsep}{4pt} 
\begin{tabular}{|wc{7em}|wc{7em}|wc{4em}|wc{4.5em} | wc{2.8em}|wc{3.4em}|wc{2.3em}|wc{4.1em}|}
\hline
\multirow{2}{*}{\textbf{Dataset}} & \multirow{2}{*}{\textbf{Specialty}} & \multirow{2}{*}{\makecell{\textbf{Num. of} \\ \textbf{Questions}}}
 & \multirow{2}{*}{\makecell{\textbf{General} \\ \textbf{Perception}}}
 & \multicolumn{4}{c|}{\textbf{Endoscopic Complexity}} \\ \cline{5-8}

 & & & & Simple & Medium & Hard & Very Hard \\ \hline
HeiCo-VQA-Base  & Rectal Surgery & 167,384 & \checkmark & - & - & - & - \\ \hline
SSG-VQA & Cholecystectomy & 883,254 & - & \checkmark & \checkmark & \checkmark & - \\ \hline
Kvasir-VQA & Endoscopy & 58,798 & - & \checkmark & \checkmark & \checkmark & - \\ \hline
Endoscapes-CVS & Cholecystectomy & 7,643 & - & - & - & - & \checkmark \\ \hline
\end{tabular}

\label{tab:endoscopic_complexity}
\end{table}

To estimate performance on more \textbf{complex tasks} requiring surgical knowledge and advanced reasoning, we convert existing endoscopic vision tasks into VQA formats. Classification tasks are straightforwardly rephrased into multiple-choice questions, enabling evaluation of nuanced surgical decision-making. Our framework seamlessly integrates any model that works with Huggingface Transformers \cite{transformers_wolf} or is accessible via APIs, ensuring broad compatibility with SOTA VLMs. Performance is evaluated using two primary metrics. The \textbf{Accuracy(\%)} metric \cite{bridging_raedsch} assesses model performance on image–question pairs by calculating the fraction of correctly answered questions per image; an image is deemed successful if this fraction meets or exceeds a specified threshold (e.g., 75\%). Hence, a metric value of 0.5 at threshold 0.75 means that in 50\% of images at least 75\% of questions have been answered correctly. Additionally, \textbf{Matthew’s Correlation Coefficient} provides a balanced measure of classification performance, a value of zero indicates performance equivalent to a random classifier.

\begin{figure}[t]
\includegraphics[width=\textwidth]{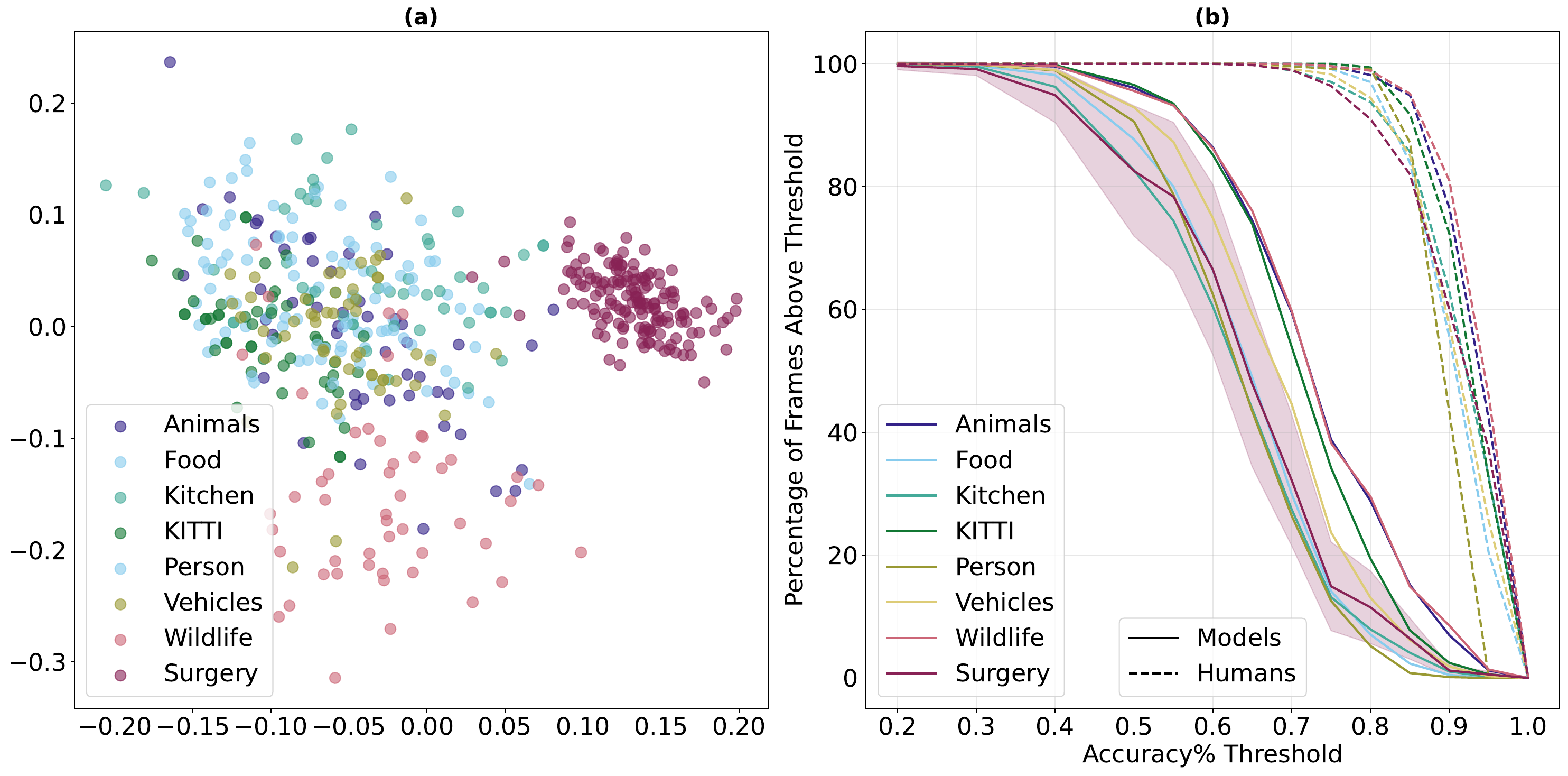}
\caption{\textbf{Basic perception tasks are solved with comparable accuracy on surgical and natural images.} (a) PCA Visualization of the Qwen2 VL 7B vision encoder showing that surgical images (red) are encoded differently from natural ones. (b) Despite this, SOTA models perform similarly when solving basic perception tasks on the different domains. Each curve represents the average performance across models according to the Accuracy\%(t) metric, which provides the percentage of images for which at least a specified proportion of questions (on the y-axis) are correctly answered. For the surgery data, bands represent the standard deviation (±) across the models. Performance of a random classifier is dataset-dependent, but generally close to zero.} 
\label{figure:fig3}
\end{figure}

\subsection{New Dataset HeiCo-VQA-Base} \label{section:sec22}
We created a new surgical VQA benchmarking dataset for basic visual perception tasks by applying our framework to the existing Heidelberg Colorectal (HeiCo) dataset \cite{heico_maier-hein}. HeiCo comprises 30 laparoscopic surgical videos (10 each from proctocolectomy, rectal resection, and sigmoid resection procedures) with 10,040 frames containing instance segmentations. For each frame, we generated an average of 17 questions across all 7 categories depicted in Fig. \ref{figure:fig2}, resulting in 167,384 question-answer pairs. To establish a human baseline, we obtained reference answers (3 to 6 with early stopping based on human agreement) through annotations outsourced to QualityMatch GmbH, Heidelberg. In total, human reference annotations were obtained for 15,844 question-answer pairs.

\begin{figure}[t]
\includegraphics[width=\textwidth]{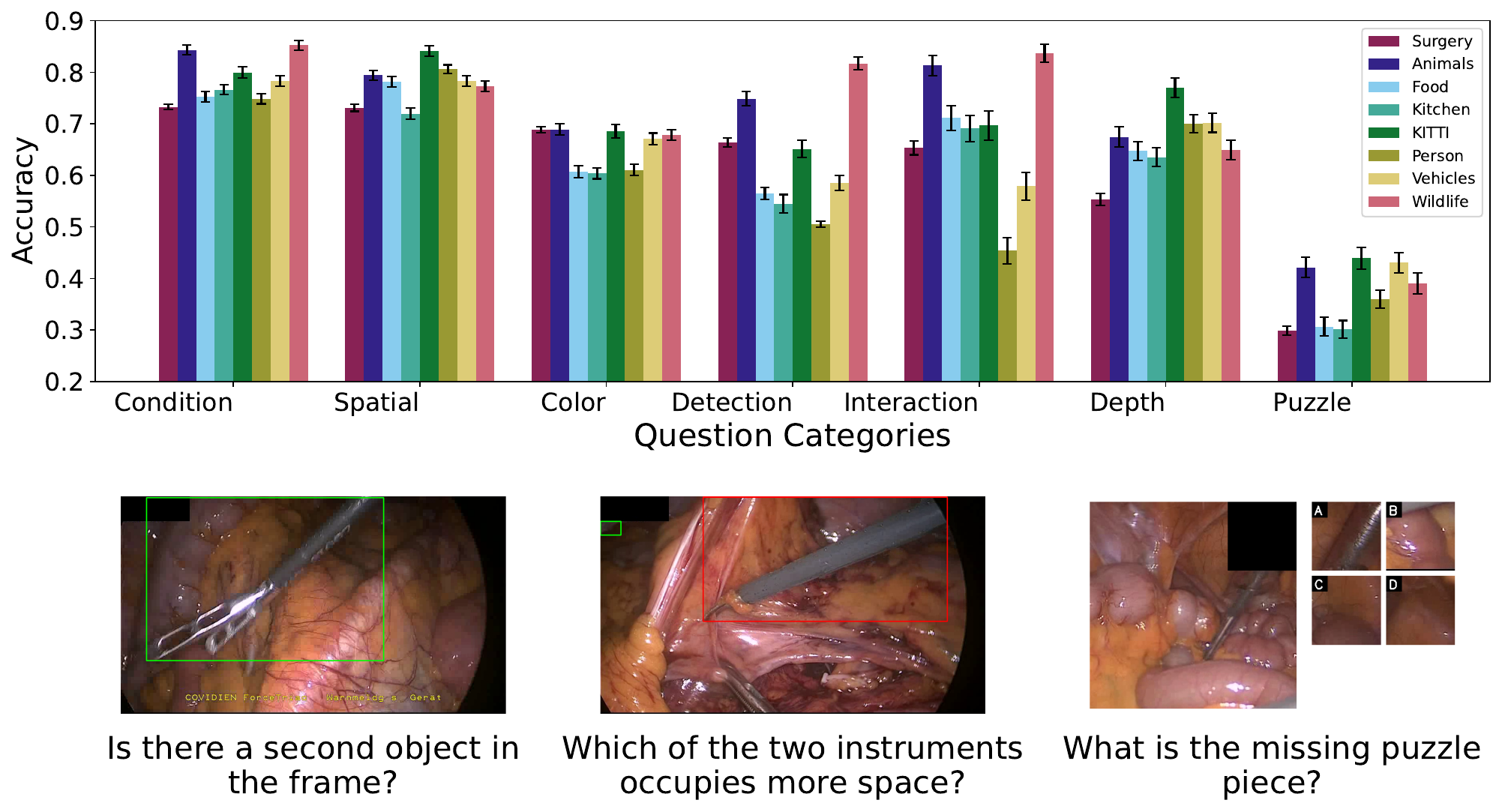}
\caption{\textbf{Vision-Language models struggle on similar tasks in surgery as in other domains.} Performance on various task categories summarized in Fig. 2 and representative sample questions. Models struggle especially with the puzzle question on the right. The error bars of each individual bar, correspond to the 95\% confidence intervals that were calculated by performing bootstrapping for each question category using 10,000 resamplings.
} \label{figure:fig4}
\end{figure}

\subsection{Experimental Setup} \label{section:sec23}
\paragraph{RQ1: Basic Perception Tasks on Surgical Images}\hfill \break
To answer RQ1, we compared the performance achieved by SOTA models on natural domains to that obtained for our new dataset. Specifically, we applied our framework to the eight domains depicted in Fig. \ref{figure:fig3}. Note that the questions proposed in prior work \cite{bridging_raedsch} were filtered by removing questions that cannot reasonably be applied to the surgical domain (e.g. determining whether an image is rotated). 
A total of 6 state-of-the-art models identified in \cite{bridging_raedsch} (GPT-4o, Claude 3.5 Sonnet, Gemini 1.5 Pro, Gemini 1.5 Flash, Qwen2-VL 72B and Llama 3.2 90B) were applied by prompting them with the question along with the image.

\begin{figure}[h]
\includegraphics[width=\textwidth]{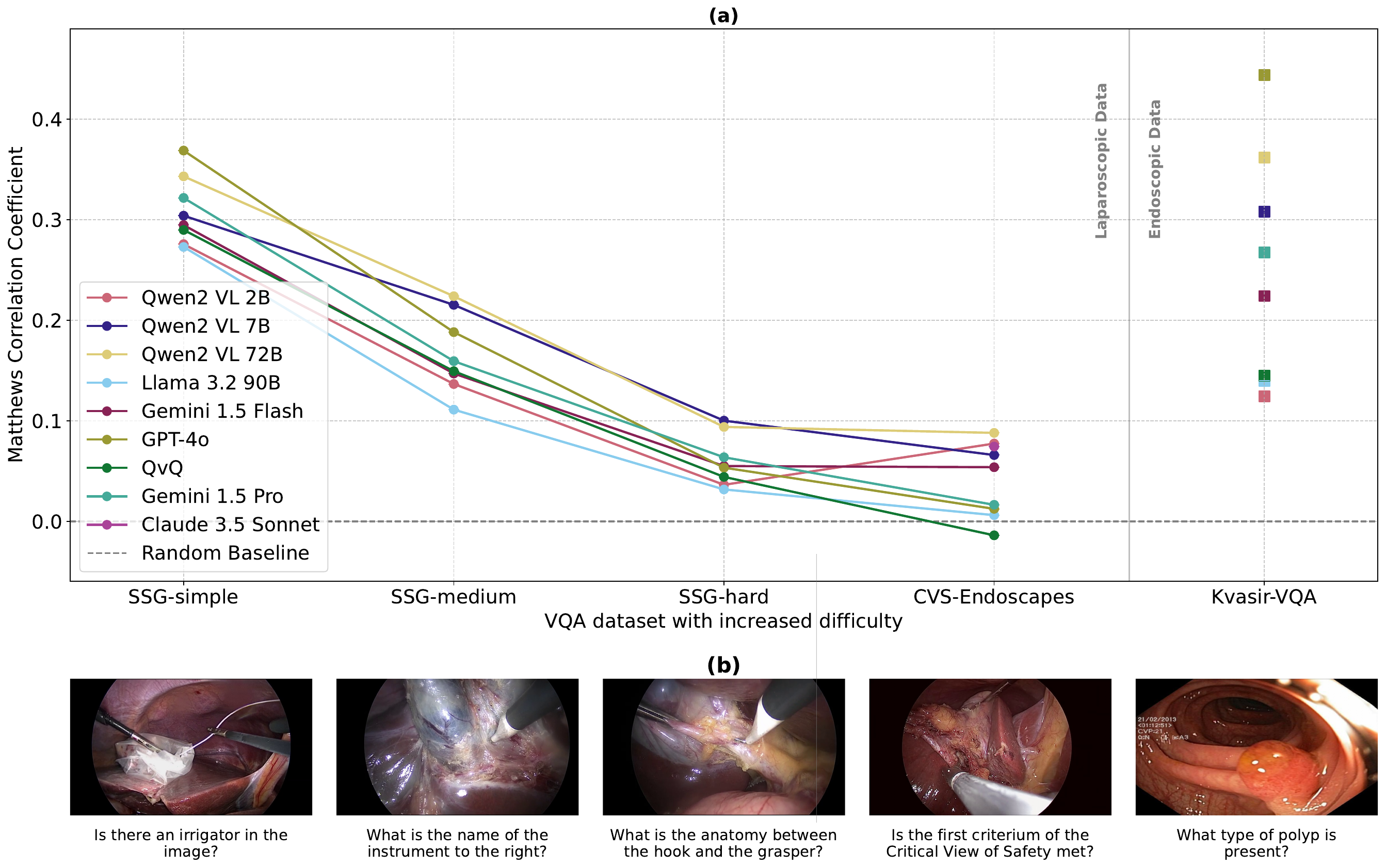}
\caption{\textbf{(a) Matthews Correlation Coefficient (MCC) is depicted as the function of task complexity for all advanced tasks on laparoscopic data (left)and separately for endoscopic data (right). (b) Example questions rephrased for brevity.}} 
\label{figure:fig5}
\end{figure}

\paragraph{RQ2: Advanced Frame-based Visual Scene Understanding}\hfill \break
We answered RQ2 with three different datasets (see Tab. \ref{tab:endoscopic_complexity}), for which we generated/adopted endoscopic vision understanding tasks of different level of complexity. Sample questions are depicted in Fig. \ref{figure:fig5}.
SSG-VQA \cite{ssgvqa_yuan} builds upon the existing Cholec80 \cite{Twinanda2016EndoNetAD} dataset, containing questions about objects, their attributes, and inter-object relationships. Some of the questions are structured across three complexity levels—simple, medium, and hard (called \textit{zero-hop}, \textit{one-hop} and \textit{single-and} in \cite{ssgvqa_yuan})—each demanding progressively more sophisticated visual reasoning capabilities. We randomly sampled 500 frames, questions and corresponding answers from those question-answer pairs that came with complexity ratings.
We leveraged Endoscapes-CVS \cite{endoscapes_murali} which is annotated with ratings for the three criteria for Critical View of Safety (CVS) for a clinically relevant highly complex task. Specifically, 1,000 questions frames were randomly sampled  and converted into questions for VQA benchmarking.
Kvasir-VQA \cite{kvasirvqa_gautam} contains questions on endoscopic images, encompassing both gastroscopic and colonoscopic examinations. Each image in the dataset is annotated with answers to six different types of questions, providing comprehensive coverage of clinical visual understanding tasks. We randomly selected 2,000 frames for benchmarking, while removing questions with multiple possible correct answers.

For VQA benchmarking, we expanded the set of  SOTA models used to solve basic perception tasks in order to investigate scaling and reasoning behavior more deeply. Specifically, we added the 2B and 7B variants of Qwen2 VL \cite{qwen2vl_wang}, which share an identical 675M-parameter vision encoder with the 72B variant. We also included QVQ 72B, an early open-weight reasoning VLM built on Qwen2 VL 72B, to assess the impact of chain-of-thought reasoning capabilities. 

We maintained consistent conditions across all evaluations by setting temperature to 0, as is customary in VLM benchmarking frameworks. For advanced tasks, the list of possible answers was appended to the prompt. All responses underwent basic post-processing answer cleaning (e.g. removal of thinking tokens) to ensure standardized evaluation.

\paragraph{RQ3: Comparison of Specialized Medical VLMs versus Generalist Models}\hfill \break
In RQ3, we explicitly compared the performance of medically fine-tuned VLMs and generalist VLMs for complex tasks by choosing pairs of models with matched size and architecture.
Specifically, we focused on the latest model family Mini-InternVL \cite{miniinternvl_gao}, which offers both generalist and medically tuned versions across three scales (1B, 2B, and 4B) for a direct comparison on surgical tasks.

\section{Results}
\paragraph{RQ1: Basic Perception Tasks on Surgical Images}\hfill \break
Endoscopic vision data typically occupies a distinct encoding space compared to natural images \cite{fingerprinting_godau}, as demonstrated with the PCA visualization of the Qwen2 VL 7B vision encoder) (Fig. \ref{figure:fig3}a). Although VLMs are primarily trained on natural images, our study reveals that they process surgical images as effectively as natural images for basic perception tasks (Fig. \ref{figure:fig3}b). All models perform substantially worse than humans across all domains. The challenges encountered by VLMs when processing surgical images were found to be similar in nature to those faced when processing images from other domains (Fig. \ref{figure:fig4}).

\paragraph{RQ2: Advanced Frame-based Visual Scene Understanding}\hfill \break
Model performance consistently decreased as task complexity increased across all model variants (Fig. \ref{figure:fig5}). On SSG-hard and CVS, all models achieved only marginally better results than random chance. Looking especially at the Qwen family, larger models (7B and 72B) consistently outperformed the smaller 2B variant. Notably, the QvQ model, despite being specifically designed for reasoning tasks, performed worse than both 7B and 72B models across all evaluated tasks. Testing on the Kvasir-VQA dataset confirmed these findings, demonstrating no performance improvement from reasoning-enhanced models when handling complex questions.

\begin{figure}[h]
\includegraphics[width=\textwidth]{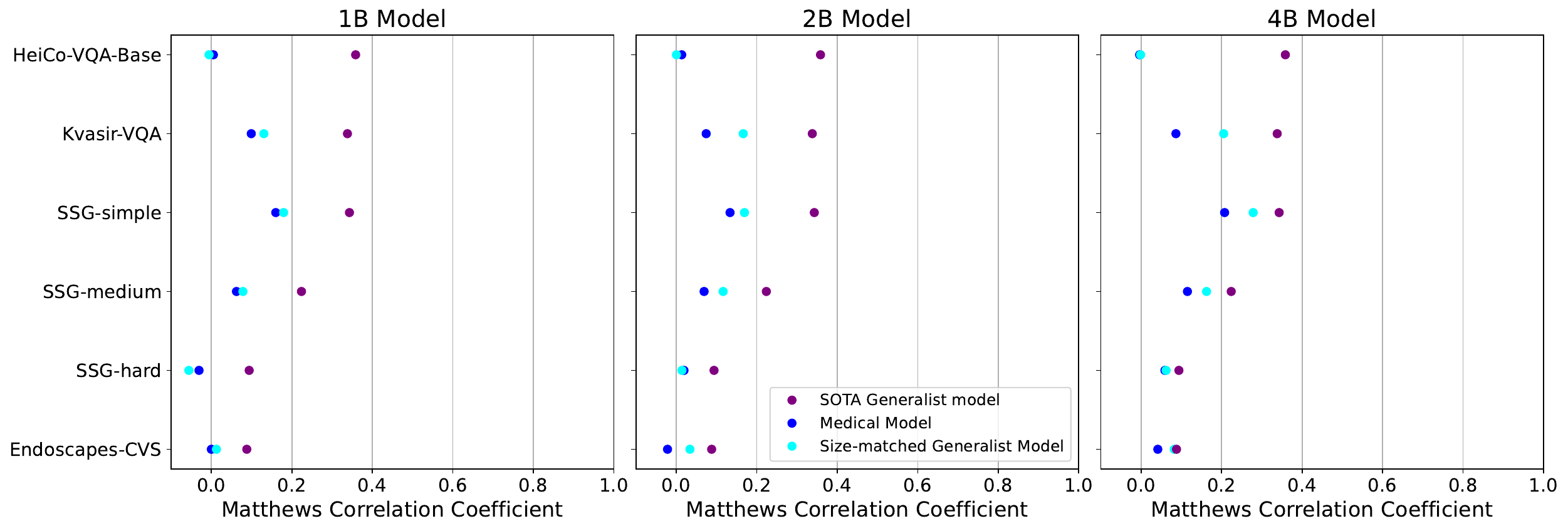}
\caption{\textbf{Medically fine-tuned models fail to outperform matched generalist models on surgical questions.} For all advanced tasks, Matthews Correlation Coefficient (MCC) is depicted for the medical model, a size-matched generalist and the state-of-the-art generalist Qwen2VL-72B.} \label{figure:fig6}
\end{figure}

\paragraph{RQ3: Comparison of Specialized Medical VLMs versus Generalist Models}\hfill \break
Medical foundation models consistently underperformed compared to generalist vision-language models (VLMs) across all evaluation metrics (Fig. \ref{figure:fig6}). This performance gap remained evident throughout all datasets and across varying levels of task complexity. Notably, neither specialized medical models nor generalist models achieved performance comparable to Qwen2VL-72B.

\section{Discussion}
While our paper does not present a new method for solving a particular (class of) problem, it provides important insights to guide further research in the endoscopic vision community through the first evaluation of zero-shot question-answering VLMs.

Our findings show that generalist VLMs perform comparably in surgery as they do in other domains when addressing similar questions. Hence, rather than waiting for specialized models, researchers should be more courageous applying generalist models to their surgical data science problems. Given that their basic image understanding is already quite strong and will likely continue to improve with new base models, the key focus should be on how to inject surgical knowledge effectively. The performance gap we observed between medical and generalist models is understandable, as medical models like the medical adaption of Mini-InternVL were trained on databases with limited surgical content.

Our analysis was limited to static images, as many accessible state-of-the-art VLMs don't yet process video input. Benchmarking questions should continue to evolve, as current versions may not fully capture the real-world complexity of surgery, and the clinical relevance of model performance may vary depending on the use case. Future research should explore surgical-specific adaptations of VLMs beyond fine-tuning, including optimized prompt design with in-context examples to better leverage surgical context. 
While evaluations of VQA models exist in various medical disciplines, ours is the first large-scale study focused on surgery and endoscopy. This is partly because early VLMs in surgery were limited to CLIP-based methods, such as HecVL \cite{hecvl_yuan}, which lacked open-ended visual question-answering capabilities.

In conclusion, our study provides key insights into both the strengths and limitations of current VLMs in surgical applications. Our findings highlight the need to move beyond basic image encoding improvements and toward strategies that effectively integrate surgical expertise into large generalist models.

\newpage

\section*{Acknowledgements}
We gratefully acknowledge support from the Surgical AI Hub Germany.
T. Rädsch was supported by a scholarship from the Hanns Seidel Foundation with funds from the Federal Ministry of Education and Research Germany (BMBF).
We thank the authors of the HeiCo dataset, especially Martin Wagner, for releasing this valuable dataset.

\bibliographystyle{elsarticle-num}

\bibliography{lib}

\end{document}